%% file: example_paper.tex
\documentclass{article}

\usepackage[T1]{fontenc}

\usepackage{microtype}
\usepackage{graphicx}
\usepackage{subcaption}
\usepackage{booktabs} 
\usepackage{longtable} 
\usepackage{listings}  
\usepackage{multicol}  
\usepackage{enumitem}

\usepackage{hyperref}

\usepackage[accepted]{icml2026}

\makeatletter
\renewcommand{\Notice@String}{Accepted by the ICML 2026 AI for Science Workshop.}
\makeatother

\usepackage{amsmath}
\usepackage{amssymb}
\usepackage{mathtools}
\usepackage{amsthm}

\usepackage[capitalize,noabbrev]{cleveref}

\theoremstyle{plain}

\theoremstyle{definition}

\theoremstyle{remark}

\usepackage[textsize=tiny]{todonotes}

\icmltitlerunning{An AI Scientist that Doesn't Drift}

\begin{document}

\twocolumn[
  \icmltitle{An AI Scientist that Doesn't Drift: Taste, Structure, and Falsifiable Findings in a Quadruped Navigation Research Loop}



  \icmlsetsymbol{equal}{*}

  \begin{icmlauthorlist}
    \icmlauthor{Yiwen Zhang}{equal,pma}
    \icmlauthor{Eloise Zeng}{equal,mce}
    \icmlauthor{Jaeha Lee}{equal,pma}
    \icmlauthor{Tony Yue Yu}{pma}
  \end{icmlauthorlist}

  \icmlaffiliation{mce}{Department of Mechanical and Civil Engineering, California Institute of Technology, Pasadena, CA 91125, U.S.A.}
  \icmlaffiliation{pma}{Division of Physics, Mathematics and Astronomy, California Institute of Technology, Pasadena, CA 91125, U.S.A.}

  \icmlcorrespondingauthor{Jaeha Lee}{jaeha@caltech.edu}
  \icmlcorrespondingauthor{Tony Yue Yu}{yuyuetony@gmail.com}

  \icmlkeywords{Machine Learning, ICML}

  \vskip 0.3in
]



\printAffiliationsAndNotice{\icmlEqualContribution}

\begin{abstract}

Autonomous research loops driven by large language models can run machine-learning experiments at scale but tend to drift toward local refinements of whichever metric they optimise rather than testing the hypotheses that motivate the experiments. We address this structurally and present an AI Scientist for studying generalisation in quadruped robot navigation policies in simulation. Building on the autoresearch paradigm of Karpathy, our loop adds three components: an immutable \emph{experiment card} that pairs each iteration's prediction with its outcome under a fixed schema, so a falsified hypothesis cannot be retconned; specialised subagents restricted to mechanical roles; and \textbf{kkanbu}, a preference oracle that holds the user's research taste as a typed knowledge graph and is the only component permitted to make subjective judgements. To isolate the oracle we run the identical loop twice across eleven research streams, with and without kkanbu. Neither arm drifts: both falsify roughly three quarters of their own hypotheses, and the best trained policy comes from the oracle-less arm. What the oracle changes is direction, not score: it alone explores test-time adaptation, it authored the winning designs where its arm led, and it carried lessons across streams that the other arm repeatedly re-derived. The scaffold keeps the loop honest; kkanbu decides where it looks.

\end{abstract}

\section{Introduction}

Learned navigation policies are increasingly studied as alternatives or complements to classical path-planning pipelines. In place of hand-engineered planners, neural policies can produce more reactive behaviors and adapt to edge cases. Quadruped navigation is harder: the policy must coordinate obstacle avoidance with feasible locomotion. Generalization is also unsettled. A policy trained on one distribution typically degrades when conditions shift, and which training-time choices close that gap is not known.

We study a hierarchical setup for the Unitree Go1 quadruped, where a frozen locomotion policy executes motion commands $(v_x, v_y, v_{yaw})$ predicted by a learned navigation policy from LiDAR and goal observations. Policies train in procedurally generated obstacle environments and are evaluated across rising obstacle densities to measure out-of-distribution (OOD) generalization.

Each candidate paradigm has known tradeoffs. Imitation learning is stable and data-efficient but inherits the expert's blind spots~\cite{pomerleau1989alvinn,codevilla2018end,bojarski2016end}. Online reinforcement learning can find adaptive avoidance behaviors past the demonstrations, but sparse navigation rewards and collision penalties make optimization hard~\cite{mnih2015human,lillicrap2015continuous,schulman2017ppo}. Offline RL is a middle ground whose performance depends on dataset coverage and action-space support~\cite{fujimoto2021minimalist,kostrikov2021offline}. Which of these generalize best under density shift is open.

The design space then has many interacting choices: motion-command versus waypoint prediction, how much action history to observe, how to shape rewards against specific failure modes, and whether to evaluate paradigms in isolation or as hybrids. That structure suits autonomous research loops. A useful AI Scientist should not just run experiments at scale; it should keep unresolved hypotheses, negative results, and proposed failure mechanisms alive across batches, so later experiments test open questions instead of polishing the current leader.

We therefore built an AI Scientist system for quadruped navigation. An earlier autoresearch-style loop took a markdown problem statement and autonomously proposed, implemented, and evaluated experiments~\cite{karpathy2026autoresearch}. In practice, its choices drifted toward refinements of the current best configuration. The new system adds \textbf{kkanbu}, an oracle that holds the researcher's taste as a queryable knowledge graph, together with a structured multi-agent workflow for planning, implementation, debugging, evaluation, and analysis. Structured experiment records capture pre-experiment predictions, outcomes, and post-hoc interpretation, so findings and failed hypotheses persist across batches rather than vanishing into chronological logs.

We instantiate the system on a shared obstacle-navigation benchmark spanning privileged imitation learning, waypoint-based imitation learning with Pure Pursuit control, offline RL, and online RL, and we run it twice: once with kkanbu at the four direction-setting hand-offs, and once with no taste artifact at all. Both arms behave like researchers rather than leaderboard climbers: they falsify most of their own hypotheses, honour pre-registered falsification thresholds against the pull of the score, and converge on mechanism-level explanations. The best trained policy comes from the oracle-less arm; the oracle arm alone explores test-time adaptation and produces the highest absolute result on frozen weights.

Our contributions are threefold.\footnote{Code, the full experiment records of both ablation arms, and the frozen oracle-profile snapshot are available at \url{https://github.com/Jaeha0526/autoresearch_with_kkanbu}.} First, we introduce an AI Scientist architecture that separates structural honesty (immutable experiment cards, content-neutral subagents) from research direction (a queryable taste oracle). Second, we report a controlled two-arm ablation of the oracle across eleven research streams: structure alone eliminates the drift pathology, and the oracle's measurable contribution is breadth across the user's search axes, authorship of specific winning designs, and cross-stream memory, at no score premium. Third, we present mechanism-level findings from the runs, including a representation boundary jointly exposed by the two arms and a deployable test-time filter invented by the oracle arm.


\section{Related Work}

\subsection{Learned Navigation for Robots}

Learned navigation policies have long served as alternatives or complements to classical planning for mobile robots~\cite{pomerleau1989alvinn,bojarski2016end,sadeghi2016cad2rl}, spanning reinforcement learning, imitation learning, and hybrid methods on wheeled robots, drones, and quadrupeds~\cite{kahn2018self,tai2017virtual,chiang2019learning}. Hierarchical control is standard for legged systems: a low-level locomotion controller executes commands from a higher-level navigation policy~\cite{hwangbo2019learning,lee2020learning,liang2024navigation}. Privileged imitation learning gives stable supervision from expert planners; reinforcement learning can discover adaptive behaviors beyond the demonstrations~\cite{ross2011dagger,codevilla2018end}.

Relative to wheeled robots and drones, systematic study of obstacle-density generalization in quadruped navigation remains limited. We address this through an autonomous AI Scientist workflow that proposes, evaluates, and revisits navigation hypotheses across experiment batches.

\subsection{Autonomous Research Agents and AI Scientists}

Recent AI Scientist systems use large language models to propose experiments, modify code, launch training jobs, analyze results, and write reports. \emph{autoresearch}~\cite{karpathy2026autoresearch} runs an agent in a tight loop: given a program description and evaluation metric, the agent proposes a code change, trains briefly, and keeps the change only if the metric improves. \emph{The AI Scientist}~\cite{ai_scientist_2024} generates and evaluates research ideas; \emph{Co-Scientist}~\cite{co_scientist_2025} uses multi-agent workflows to generate and critique hypotheses; Voyager~\cite{wang2024voyager} pursues long-horizon autonomous learning through tool use.

These systems handle the execution side of research well: proposing variations, running jobs, editing code. The harder question is what to run next, which directions to abandon, and which results merit deeper investigation. Such judgements of research direction are not derivable from the metric being optimised, and without an external source of direction an autoresearch loop tends to converge on local refinements of the current best configuration. Our work targets this gap. The proposed system preserves unresolved hypotheses, failed explanations, and prior findings across batches in structured experiment records, and routes every direction-setting decision through \textbf{kkanbu}, a queryable oracle holding one researcher's taste. We measure what the oracle adds by running the loop with and without it.

\section{Methods}
\label{sec:methods}

We compare five paradigm streams on a forest-navigation benchmark: PIL Control (Section~\ref{sec:pil_control}), PIL Trajectory (Section~\ref{sec:pil_trajectory}), offline RL with IQL (Section~\ref{sec:iql}), and online RL with PPO, in pure and privileged-critic variants (Section~\ref{sec:ppo}).

\subsection{Problem Setup}
\label{sec:problem_setup}

All experiments are in simulation; we make no sim-to-real claim. Our focus is the training-time choices that govern out-of-distribution generalisation, and the autoresearch loop is more tractable in simulation.

A Unitree Go1 navigates from a randomly initialised spawn to an $(x,y)$ goal in a $30\,\mathrm{m}$ square room. Cylindrical obstacles are sampled from a Poisson point process with intensity $\delta$ (default $\delta = 0.04\,\mathrm{m}^{-2}$), simulated in MuJoCo~\cite{todorov2012mujoco} via MJX~\cite{freeman2021brax,zakka2025playground}. The 75-D observation combines 64 LiDAR beams with a goal encoding and a velocity-command history; actions are velocity commands at $50\,\mathrm{Hz}$. Full environment, observation, and action specifications are in Appendix~\ref{app:implementation}.

We evaluate across $\delta \in \{0.01,\allowbreak 0.02,\allowbreak 0.04,\allowbreak 0.08,\allowbreak 0.12,\allowbreak 0.16\}\,\mathrm{m}^{-2}$, with 100 episodes per density. To emphasise robustness in clutter, we report a weighted composite
$S_{\mathrm{comp}} = (1.0\,S_{0.01} + 1.0\,S_{0.02} + 1.0\,S_{0.04} + 1.5\,S_{0.08} + 2.0\,S_{0.12} + 2.5\,S_{0.16}) / 9.0$,
where $S_{\delta}$ is success rate at density $\delta$.

Density is the primary axis of distribution shift. Preliminary sweeps over obstacle shape, room size, and obstacle size produced single-digit OOD drops; density produced tens of points at the densest setting, so we concentrate on it.

\subsection{Hierarchical Control Architecture}
\label{sec:hierarchical}

A frozen locomotion policy~\cite{schulman2017ppo,lee2020locomotion,miki2022perceptive}, pre-trained for $10^8$ steps, maps proprioception and the velocity command to 12-D joint targets, isolating \emph{navigation} as the variable under study. The trainable navigation layer outputs the velocity command, scaled by $v^{\max}$ before the locomotion controller.

\subsection{Privileged Imitation Learning (PIL) Control}
\label{sec:pil_control}

PIL Control trains a student to imitate a privileged expert. The expert sees ground-truth environment information; the student sees only onboard observations.

\paragraph{Expert.}
A* on a $0.2\,\mathrm{m}$ occupancy grid plans paths, which a Pure Pursuit~\cite{coulter1992purepursuit} tracker follows. We choose A*~\cite{hart1968astar} over RRT/RRT*/PRM* because it is complete and optimal on the grid; sampling-based planners are only asymptotically optimal and produce nondeterministic teacher trajectories~\cite{karaman2011sampling}. Pure Pursuit uses proportional yaw control ($k_p = 1$), an alignment-modulated forward speed $\max(\cos \theta_{\mathrm{err}}, \mathrm{MAF})$, and a slowdown $\min(d/d_{\mathrm{slow}}, 1)$ with $d_{\mathrm{slow}} = 2\,\mathrm{m}$.

The main knob is the minimum alignment factor $\mathrm{MAF}$. $\mathrm{MAF}=0.3$ produces smooth always-forward demos; $\mathrm{MAF}=0.0$ allows stop-and-rotate, yielding a bimodal command distribution (\textbf{MAF0}).

\paragraph{Student.}
A 3-layer MLP with hidden dims $(512,256,128)$ and Swish activations~\cite{ramachandran2017swish} maps 75-D observations to actions $a^* = (v_x, v_y, v_{\mathrm{yaw}})$ via tanh-rescaled outputs. Training uses $\sim\!10^7$ expert transitions under MSE on velocities (Equation~\ref{eq:pil_loss}); Adam with cosine-annealed LR $10^{-3}$, 1k warmup steps, patience-based early stopping. Remaining hyperparameters are in Appendix~\ref{app:hyperparams}.

\subsection{PIL Trajectory}
\label{sec:pil_trajectory}
\paragraph{Why waypoints, not velocities?} In clutter, similar observations admit multiple valid avoidance actions (left vs.\ right around the same obstacle). Direct behaviour cloning (BC) on velocities therefore gives ambiguous supervision; under MSE the model averages modes and oscillates near obstacles. Waypoint prediction supervises a coherent future path: geometrically inconsistent trajectories incur large coordinate-space error, forcing the network to commit to one strategy. Recent navigation systems use the same trick.

The student maps $o\!\in\!\mathbb{R}^{75}$ to $\tau = s\,\tanh(f_\theta^\mathrm{raw}(o))\!\in\!\mathbb{R}^{2H}$ with $H\!=\!10$, $\Delta t\!=\!0.5$\,s, $s\!=\!5$\,m, trained with MSE on the predicted waypoint sequence (Equation~\ref{eq:traj_loss}).

\paragraph{LiDAR-Attention head.} A flat MLP discards LiDAR's bearing structure. We treat each beam as a token, add a positional code $p_i\!=\!i/64$, run 4-head self-attention~\cite{vaswani2017attention} over the 64 tokens, and pool with a learned per-token gate $\alpha_i\!\in\![0,1]$ to obtain $z_\mathrm{lidar}=\sum_i\alpha_i\tilde h_i$. The fused $[z_\mathrm{goal}\,\|\,z_\mathrm{lidar}\,\|\,\alpha]\!\in\!\mathbb{R}^{192}$ feeds a $(256,128)$ head. Attention is preferable to PointNet-style max-pool~\cite{qi2017pointnet}: it is a soft, content-dependent aggregator that adapts between sparse-forest (a few critical beams) and dense-forest (distributed signal) regimes, and the exposed $\alpha$ is an interpretable diagnostic~\cite{patel2024lidarattention}.

\paragraph{Inference-time tracking.} Predicted waypoints feed the same Pure Pursuit controller as the expert (Section~\ref{sec:pil_control}); look-ahead index 5.

\subsection{Implicit Q-Learning (IQL)}
\label{sec:iql}

IQL~\cite{kostrikov2022iql} avoids OOD action queries by replacing the policy-improvement max with an in-sample expectile (Equations~\ref{eq:iql_v},~\ref{eq:iql_tau}) at $\tau\!=\!0.7$. The Q-head trains via standard TD against $V_\psi$ with double-Q clipping. The policy is extracted by advantage-weighted regression (AWR; Equation~\ref{eq:iql_awr}) with $\beta\!=\!3$, $c\!=\!5$ (Kostrikov MuJoCo defaults). We choose IQL for its small tuning surface (no Q-ensemble, no behaviour model), $\sim\!4\times$ speedup over CQL~\cite{kumar2020cql} on D4RL~\cite{fu2020d4rl}, and because not-querying-OOD directly fits our expert-only dataset (no negatives). The dataset is the same $\sim\!8.65\!\times\!10^6$ A*-Pure-Pursuit transitions as PIL; the labelled reward sums a sparse goal bonus, a dense distance-reduction term, and a path-following term. Critic and value heads are $3\!\times\!256$ MLPs; the actor is $2\!\times\!256$. We also evaluate \textbf{IQL-Attention}, in which the three networks share the LiDAR-Attention extractor of Section~\ref{sec:pil_trajectory}.

%
%
\subsection{Proximal Policy Optimization (PPO)}
\label{sec:ppo}

We evaluate PPO~\cite{schulman2017ppo} as the online-RL baseline. In the standard setting, both actor and critic see only the 75-D onboard observation, and the policy outputs $(v_x, v_y, v_{\mathrm{yaw}})$.

We additionally evaluate a privileged-critic variant: during training the critic sees A* waypoints from start to goal on the $0.2\,\mathrm{m}$ grid, post-processed into a shortest-path trajectory, while the actor still observes only the onboard input. The privileged setup adds a reward for progress toward the upcoming waypoint.

The initial reward (both variants) was hand-designed to encourage goal progress and completion while penalising collisions, falls, and abrupt velocity changes. Subsequent autoresearch iterations explored backward-velocity, time-to-collision (TTC), and stuck-detection penalties.

\subsection{Anchor density-generalization profile}
\label{sec:baselines}

\begin{figure}[t]
  \centering
  \includegraphics[width=\columnwidth]{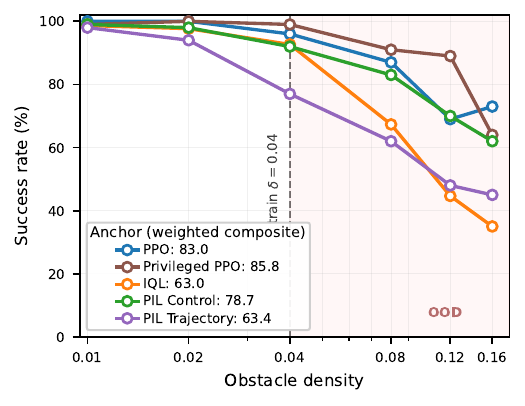}
  \caption{Per-paradigm anchor profiles under obstacle-density
  shift. Each curve is the stream's anchor: the human-seeded
  baseline (or default-configuration first batch) that every later
  experiment in that stream is scored against. The legend gives
  each anchor's OOD-weighted composite. Training density
  $\delta=0.04$ is dashed; the OOD region ($\delta\!>\!0.04$) is
  shaded.}
  \label{fig:baselines}
\end{figure}

Before describing the AI Scientist system, we fix the baselines it
must beat. Each of the five paradigm streams starts from an
\emph{anchor}: the best manually-tuned policy for that paradigm where
one existed, otherwise the stream's default configuration.
Figure~\ref{fig:baselines} plots each anchor's density-vs-success
profile. All were trained at $\delta=0.04$ and evaluated across the
six-density sweep $\{0.01, 0.02, 0.04, 0.08, 0.12, 0.16\}$, and every
improvement claim in Section~\ref{sec:ablation} is a delta against
its stream's anchor composite.

The anchor profile already exposes the design space. The two
online-RL anchors sit at the top (privileged PPO $85.83$, PPO
$83.0$) and degrade most gracefully into the OOD region. PIL control
follows at $78.72$. IQL ($62.98$) and PIL trajectory
($63.39$) sit twenty points lower and collapse hardest at
$\delta\!\ge\!0.12$; the trajectory anchor's collapse is a
controller-interface artefact that Section~\ref{sec:findings}
revisits. All anchors except IQL are single-seed, and that
caveat travels with every anchor delta.

These curves are not comparable systems (they differ in supervision,
dataset, and what the actor head can emit); they are the search space
the two arms operate on. Each anchor is the null hypothesis its
stream's experiments test against.

\section{The AI Scientist System}
\label{sec:system}

\begin{figure}[t]
  \centering
  \includegraphics[width=\columnwidth]{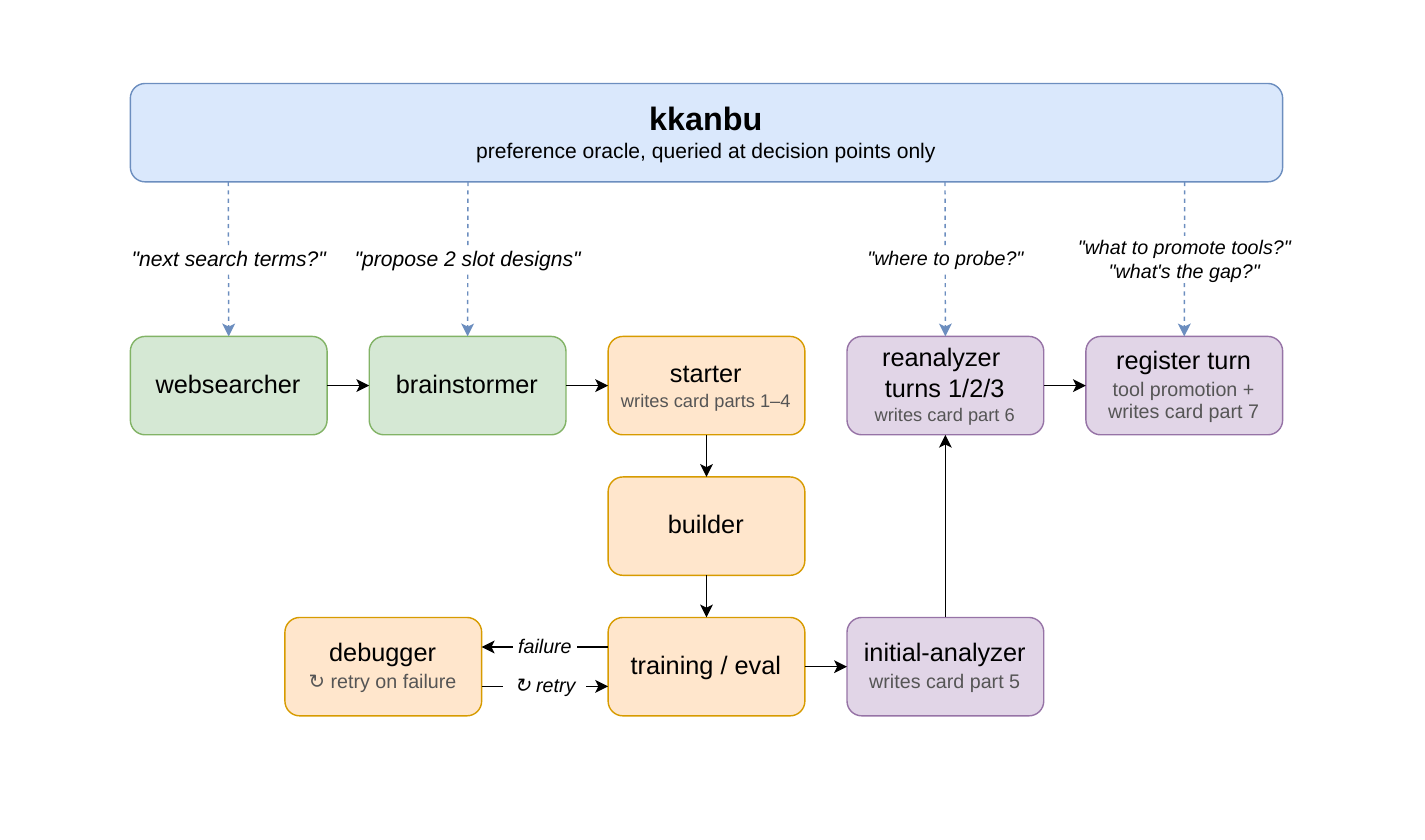}
  \caption{One stream-batch of the AI Scientist loop. \textbf{kkanbu} (blue) is the only opinion-holder, consulted at four labeled decision points. Planning agents (green) decide \emph{what to do}; execution agents (orange) carry it out; analysis agents (purple) write the card and register tools.}
  \label{fig:system}
\end{figure}

The autoresearch paradigm of \citet{karpathy2026autoresearch} is a loop in which an agent proposes a change, tests it, and keeps it only if the score improves. We extend this with eleven parallel research \emph{streams}: one per method family of Section~\ref{sec:methods}, plus six discovery streams (two hybrid, two new-paradigm, two tuning) that combine, replace, or tune the paradigm recipes. Each stream advances in serial \emph{batches} from its anchor; a batch is a single experiment with its own card, and each batch mutates a single axis off a named parent configuration.

Single-score loops are known to game the score by changing seeds, narrowing validation splits, or otherwise improving the metric without improving the underlying capability. We address this structurally, not at the prompt level, with three components: an immutable \emph{experiment card} that records each iteration's prediction and outcome under a fixed schema; specialised subagents restricted to mechanical roles, none permitted to set direction; and \textbf{kkanbu}, a preference oracle consulted at four labelled decision points and the only component permitted to make subjective judgements.

\subsection{Experiment-card Schema}
\label{sec:card}

The experiment card is the loop's central instrument: a fixed JSON schema, one card per experiment, that carries each iteration's knowledge across batches.

A card has three stages. \textbf{Planning.} \emph{Motivation} (part 2) connects the experiment to a prior finding or open question; \emph{expected result} (part 4) states the prediction whose falsification would be informative, together with an explicit falsification clause the outcome will be judged against. Both are written before training launches and become immutable once the experiment runs, so a falsified prediction cannot be retconned. \textbf{Experiment.} The card records the experiment's isolated code workspace, build notes, and the evaluation score from the per-density sweep (part 5). \textbf{Analysis.} The mechanism-analyzer separates findings from interpretation in a structured synthesis (part 6) and identifies the next open question (part 7), building purpose-built diagnostics (failure-mode classifiers, weight-trajectory probes, action-distribution analyzers); the most generalisable are promoted into the shared toolkit.

The card is the only place where a prediction (planning, immutable) and its outcome (experiment and analysis) co-reside, which is what makes the system answerable to a falsified hypothesis. The findings in Section~\ref{sec:findings} all live in the analysis stage.

\subsection{Per-batch Flow}
\label{sec:flow}

A batch advances in three phases (Figure~\ref{fig:system}) matching the card stages of Section~\ref{sec:card}.

\textbf{Planning.} A \texttt{websearcher} runs an iterative literature search. A \texttt{brainstormer} then reads the synthesis, prior cards from this stream, and curated resource notes, and produces the stream's next experiment proposal.

\textbf{Experiment.} An \texttt{experiment-starter} transcribes the proposal into the card's planning fields and creates an isolated workspace. An \texttt{experiment-builder} implements the experiment and prepares training and evaluation jobs, and a \texttt{code-reviewer} checks the build against the card and the fair-comparison envelope (Section~\ref{sec:envelope}) before anything launches. A \texttt{maintainer} walks the compute queue and updates a progress dashboard; an \texttt{experiment-debugger} applies minimal fixes to failures, appending a debug note to the card after each attempt so subsequent attempts are not blind. Each experiment runs a single seed first; two more launch only if it looks promising, concentrating compute on configurations whose preliminary signal merits replication. On success, an \texttt{experiment-initial-analyzer} writes the actual result.

\textbf{Analysis.} A \texttt{mechanism-analyzer} runs three rounds of deeper analysis, building diagnostics as needed, and synthesises findings into the card. A final register turn writes the open question and next direction and proposes generalisable diagnostics for promotion.

\subsection{What Is Pinned and What Is Free}
\label{sec:envelope}

Every experiment in every stream, in both arms, runs inside the same fair-comparison envelope, fixed in the program description that all agents read (reproduced in Appendix~\ref{app:program}). Pinned: the 64-beam LiDAR as the only perception; the frozen locomotion layer; the environment's physics, arena, and obstacle-generation logic; the training density $\delta=0.04$, held constant for a whole run (no curriculum); the six-density evaluation sweep with its composite weighting, byte-untouched; the core policy MLP (512, 256, 128); demonstration data at the same order of magnitude; and the staged three-seed protocol (one debug seed, then two replication seeds unless the first craters). Free, unless a card narrows it further: the training paradigm within the stream's scope, auxiliary modules around the core network, observation encoding and action-history length, reward shaping for the RL paradigms, controller parameters, hyperparameters, regenerating demonstration data when a design requires it, and new analysis tools. The code-review gate enforces the envelope before each launch; the post-hoc audit of Section~\ref{sec:ablation:honesty} re-checks it across all lineages.

\subsection{The kkanbu Oracle}
\label{sec:kkanbu}

\textbf{kkanbu} is a preference oracle that holds the user's research taste as a typed knowledge graph of stances, values, and reasoning patterns. Each call sends a free-text question; kkanbu retrieves the relevant subgraph and returns a response in the user's voice with a confidence score grounded in the retrieved nodes. Construction and evolution of the profile are in Appendix~\ref{app:kkanbu}.

Of the subagents in Section~\ref{sec:flow}, four may query kkanbu, one at each direction-setting hand-off (Figure~\ref{fig:system}). The \texttt{websearcher} asks for the next search terms or a stop signal. The \texttt{brainstormer} submits the gathered context and relays the resulting experiment proposal verbatim into the card. The \texttt{mechanism-analyzer}, on each of its three analysis turns, asks which metric, density, or behavioural pattern to probe next; its final register turn asks which diagnostics merit promotion and which open gap defines the next direction. Every other subagent is structurally forbidden from calling kkanbu.

One might ask why this rule is not enforced by a constraint prompt inside each subagent. A prompt encodes a static instruction; kkanbu encodes the user's taste as a graph and retrieves the subset relevant to each query. The graph captures \emph{why} the user pursues a high score on the generalisability composite of Section~\ref{sec:problem_setup}: what counts as real generalisation, which priors are persuasive, which kinds of tuning are uninteresting. It is not a list of preferred facts or technologies. When the websearcher must rank candidate term-sets, or the mechanism-analyzer must choose probes for a confusing failure, the answer follows not from a single sentence but from the user's stance on what is worth knowing.

This restriction is what lets kkanbu \emph{steer} the loop. Without it, each subagent faces local pressure toward shortcuts: a websearcher might converge on the most-cited papers, a brainstormer might prefer the experiment most likely to score well, a mechanism-analyzer might rationalise a falsified prediction. Routing every direction-setting decision through kkanbu replaces these pressures with the user's taste, and the loop can overturn its own previous-batch verdicts because kkanbu's answers are anchored to the user's stated stances, not to the recent leaderboard.

\subsection{Isolating kkanbu: a controlled ablation}
\label{sec:kkanbu_ablation}

The three components play different roles: the experiment card and the
content-neutral subagents are structural \emph{scaffolding}, while kkanbu is the
only component that supplies research \emph{direction}. Is the oracle
load-bearing, or does the scaffold alone do the same work? We run the identical
loop in two configurations to find out. In the \textbf{with-kkanbu}
configuration, the four direction-setting hand-offs of
Section~\ref{sec:kkanbu} query kkanbu live: each call retrieves a
question-specific subgraph from a profile snapshot frozen before the run and
returns a grounded, confidence-calibrated answer. In the
\textbf{without-kkanbu} configuration, the same four hand-offs are carried out
by the same subagents reasoning from the shared program description (Appendix~\ref{app:program}) alone; no
oracle and no taste artifact is consulted. The card schema, the subagent
roster, the stream structure, the demonstration data, the six-density scoring
rule, and the three-seed evaluation protocol are identical in both arms. One
property of this design matters for reading the results: the scaffold itself
already encodes the core of the user's taste (the mandatory mechanism and
falsification card fields, the OOD-weighted composite), so the comparison
measures what a live oracle adds \emph{on top of} a taste-saturated scaffold.
Section~\ref{sec:ablation} reports the comparison.


\section{Findings}
\label{sec:findings}

We present three findings from the two-arm run. Each is selected
because its lineage is fully traceable in the audit trail: the chain
of card part-7 open questions that motivated each batch. Together
they cover the three provenances the ablation makes available. The
first is a cross-arm result neither arm could have produced alone;
the second is a without-kkanbu arc showing what the scaffold alone
does; the third is a with-kkanbu arc showing what the oracle adds.

\subsection{A representation boundary, bracketed from both sides}
\label{sec:findings:representation}

The trajectory paradigm of Section~\ref{sec:pil_trajectory} predicts
waypoints that a Pure Pursuit tracker converts to velocity commands.
It was historically the project's worst performer, and both arms
started it from the same anchor (63.39). In the without-kkanbu arm,
regenerating the expert demonstrations with an evasive
strafe-and-brake layer produced that arm's champion when a
velocity-head student consumed the data ($90.96 \pm 1.52$,
Section~\ref{sec:findings:teacher}). The same class of evasive data,
forced through the waypoint representation, cratered to $8.06$ (a
single-seed run recorded as a falsified terminal; the protocol does
not require replicating clear failures): the
tracker turns toward the next waypoint and only drives forward (its
lateral velocity is hard-coded to zero, verified in source), so
demonstrations that brake and sidestep become demonstrations that
stop. The with-kkanbu arm recovered the paradigm by the complementary
route, reaching $78.48 \pm 0.92$ ($+15.1$ over the anchor): an
eval-time re-tune of the tracker's lookahead accounts for roughly
$+11$ against a matched fixed-lookahead control, and a small trained
head that adds a lateral nudge after the tracker accounts for $+4.1$
against the same control. Read together, the two arms bracket a
representation law the project had misfiled as a paradigm failure:
velocity-space outputs can express evasion, and waypoint-plus-tracker
outputs structurally cannot, whatever the data contains.

\subsection{The scaffold alone does research: a falsification chain
ends at the data generator}
\label{sec:findings:teacher}

The without-kkanbu arm's strongest result came from two falsified
batches that were mined for mechanism rather than discarded. The
first batch added a consistency loss that achieved its predicted
invariance, but the mechanism analysis showed \emph{how}: by
desensitisation. It halved the policy's LiDAR-to-yaw sensitivity,
suppressing the only evasive primitive the expert demonstrates, and
the card recorded the batch as falsified. The second batch raised
capacity and incentive and was falsified too; a student cannot fit a
response its teacher barely exhibits. The accumulated conclusion
relocated the bottleneck from the student to the teacher: the shared
expert turns but never strafes; its lateral velocity is exactly zero
across all ${\sim}8.65$ million demonstration frames. The third batch
acted on the diagnosis, regenerated the demonstrations with an
evasive strafe-and-brake layer on the expert, retrained the same
student, and produced the best trained policy of the whole comparison
($90.96 \pm 1.52$). This is the drift-resistant behaviour the
original loop lacked, produced with no oracle in the loop. The cost
of running without one shows up elsewhere: roughly seven streams
independently re-derived the same teacher diagnosis before this one
acted on it (Section~\ref{sec:ablation:oracle}).

\subsection{The oracle arm invents a deployable test-time filter}
\label{sec:findings:cone}

kkanbu framed the hybrid stream's opening experiment as a control
rather than a horse race: a coverage-versus-invariance A/B whose
\emph{control arm} became the arm's trained co-leader
($87.28 \pm 0.56$). Its next proposal was test-time adaptation, the
search axis no experiment in the without-kkanbu arm ever touched: a
collision-cone velocity filter on the frozen student, specified
verbatim in the planning transcript as a cone gate rather than a raw
distance or time-to-collision threshold, paired with a deceleration
cap. The filter reads only the LiDAR observation and the policy's own
emitted command, so it is deployable onboard. Toggling it on the same
weights moves the composite from $87.54$ to $92.43 \pm 0.69$, a
within-model $+4.89$ and the round's highest absolute number,
reported as its own class because no training occurred. One caveat is
pre-registered rather than discovered post hoc: the operating point
was chosen as the best of five gate values scored on the evaluation
densities, a selection its interval does not cover, and
Table~\ref{tab:top3_cross_round} carries that flag.

\section{With kkanbu vs.\ Without kkanbu: A Controlled Two-Arm Ablation}
\label{sec:ablation}

We ran the loop of Section~\ref{sec:system} twice under the design of
Section~\ref{sec:kkanbu_ablation}: a \emph{with-kkanbu} arm (45
experiments across the eleven streams) and a \emph{without-kkanbu} arm
(34 experiments), on the same environment, observation and action
spaces, demonstration data, per-stream anchors, scoring rule, and
three-seed protocol. Each arm was operated by a different researcher
on its own compute queue, so volume-shaped differences between the
arms (experiment counts, batch depth, pace) are environment artifacts;
the comparisons below are about decision \emph{content} only. All
composites are reported as mean $\pm$ stratified percentile-bootstrap
95\% CI over three seeds unless labelled a single-seed pilot.


\subsection{Both arms stay honest}
\label{sec:ablation:honesty}

The headline is what did \emph{not} happen. A code-level audit of all
22 stream lineages found zero violations of the fair-comparison
envelope of Section~\ref{sec:envelope} in either arm; the audit
spot-checked LiDAR-only inference, frozen locomotion, the immutable
policy core, and a byte-untouched evaluation pipeline, on both sides.
Both arms falsified roughly three quarters of their own hypotheses
(20 of 27 completed experiments in the with-kkanbu arm, 19 of 27 in
the without-kkanbu arm) and recorded the negatives as negatives. The
without-kkanbu arm honoured pre-registered falsification thresholds
against the pull of the leaderboard: one paradigm stream recorded a
$+2.6$-point improvement as \emph{falsified} because it missed the
$+3$-point bar its own card had pre-registered. The
leaderboard-chasing collapse that motivated this work did not reappear
in either arm. Structure alone, without any oracle, is sufficient to
prevent it.

\begin{figure*}[t]
  \centering
  \includegraphics[width=\textwidth]{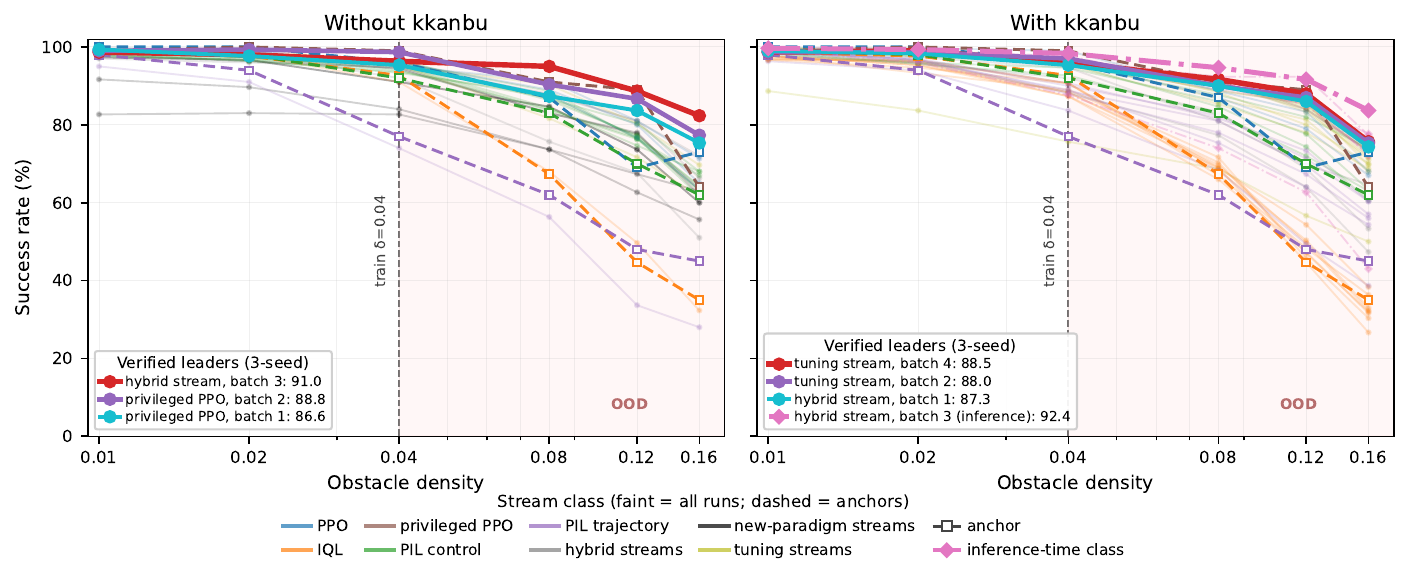}
  \caption{Density vs.\ success rate by ablation arm. Faint lines
  are all configurations with complete three-seed evaluations
  (colour-coded by stream class; legend entries name the stream and
  batch, with each entry's story in
  Table~\ref{tab:top3_cross_round}); thick lines are each arm's
  three strongest trained results; dash-dot curves in the
  with-kkanbu panel are the collision-cone filter, an inference-time
  technique on frozen weights reported as a separate class (thick:
  the reported operating point, $+4.89$ within-model; faint: its
  other evaluated gate value).
  Dashed white-fill curves are the shared per-paradigm anchors of
  Figure~\ref{fig:baselines}. Training density $\delta=0.04$ is
  dashed; the OOD region ($\delta\!>\!0.04$) is shaded. The arms
  differ in the number of evaluated configurations for environment
  reasons (different operators and queues); volume is not an oracle
  effect.}
  \label{fig:round_comparison}
\end{figure*}

\input{figures/top3_cross_round.tex}

\subsection{Scores do not separate the arms}
\label{sec:ablation:gains}

Figure~\ref{fig:round_comparison} and
Table~\ref{tab:top3_cross_round} give two views of the scoreboard,
and the honest summary is that it is close and convention-dependent.
Under the round's own acceptance convention, five streams beat their
anchors or references in the with-kkanbu arm against four in the
without-kkanbu arm; under a conservative Student-$t$ criterion the
count falls to two against three. Appendix~\ref{app:experiments}
tabulates every experiment in the frozen set, stream by stream and
batch by batch, with each design, composite, and verdict, so each
stream's full trajectory from its anchor is readable there. The without-kkanbu arm owns the best trained
result of the whole comparison ($90.96 \pm 1.52$, from regenerating
the expert demonstrations with an evasive strafe-and-brake layer
after two batches of mechanism analysis pinned the old expert as the
bottleneck) and the cleanest per-seed anchor beat ($88.80$, all three
seeds above the anchor, from a single reward-coefficient change). The
with-kkanbu arm owns a band of statistically tied trained co-leaders
($87.3$--$88.5$), the largest single-stream recovery (the trajectory
paradigm, $+15.1$ over its anchor, of which $+4.1$ is attributable to
a trained nudge head against a matched control), a categorical action
head in the control paradigm ($85.87$, $+4.5$ over a matched MSE
control), and the round's
highest absolute number: $92.43 \pm 0.69$ from a collision-cone
velocity filter applied at inference time to a frozen student,
reported as its own class because the honest attributable gain is the
within-model $+4.89$ on unchanged weights. Neither arm's advantage
survives a change of counting convention; kkanbu is not supported as
a score amplifier, and we do not claim it as one.

\subsection{What the oracle measurably changed}
\label{sec:ablation:oracle}

Three differences survive the audits, and all three are checkable in
the decision transcripts.

\textbf{Search breadth along the taste's axes.} The profile of
Appendix~\ref{app:kkanbu} names four search axes: data, inductive
bias, training objective, and test-time adaptation. The with-kkanbu
arm explored all four; the without-kkanbu arm never left the
train-a-better-policy box. Test-time adaptation appears in five
with-kkanbu experiments (including the collision-cone filter above)
and in zero of the without-kkanbu arm's 34; explicit multi-value
hypothesis sweeps appear roughly four times against once.

\textbf{Design authorship.} In the streams where the with-kkanbu arm
led, the winning levers were proposed by kkanbu verbatim in the
planning transcripts: the collision-cone filter, the trajectory
nudge head, the categorical action head above, and
the control-versus-treatment framing whose \emph{control arm} became
the trained co-leader. The authorship is symmetric in its
scepticism: handed its first confirmed win, the oracle attacked it
exactly as it attacks failures, rejecting an analyzer probe as built
to confirm itself and authoring a falsifier against its own favoured
head, which fired on a real cost the head pays at sparse density.

\textbf{Cross-stream memory and constraint policing.} The
without-kkanbu arm's streams could not see each other's negatives: it
re-tried one braking-reward lever in four experiments across three streams,
and re-derived the same diagnosis of the shared expert in seven
streams before one stream acted on it. The with-kkanbu arm moved a
measured failure from one stream into a sibling stream's design
within a day, and kkanbu repeatedly ruled proposed methods illegal
under the fair-comparison rules, including methods from its own
retrieved literature. An earlier internal run had died of exactly the
violation class this policing targets.

The ledger has a debit side, and the cards record it. kkanbu's own
paradigm bets produced the round's most expensive falsified
negatives; its quantitative predictions were routinely wrong even
when its designs won; and its hypothesis hit-rate was no better than
the scaffold's. A necessity audit of every novel idea sharpens the
debit: much of the oracle's protocol machinery was supplied ahead of
any demonstrated need and proved redundant with the scaffold's own
checks, while the without-kkanbu arm's rarer inventions, built only
when a standard tool had just failed, were load-bearing at a higher
rate. The oracle's reliable value was not idea quality. It
was choosing where the loop looked next, remembering what the loop
had already learned, and policing what the loop was allowed to try.

\subsection{What the comparison shows}
\label{sec:ablation:summary}

The scaffold keeps the loop honest; kkanbu decides where it looks.
The drift pathology is prevented by structure alone: both arms are
mechanism-chain research loops whose motivations quote measured prior
findings, because the card schema and the OOD-weighted metric encode
that discipline on both sides. What the oracle adds on top of a
taste-saturated scaffold is direction: breadth across the taste's
search axes, authorship of specific winning designs, cross-stream
memory, and live legality rulings, at no measurable score premium.
Two caveats bound these claims. Arm and operator are perfectly
confounded, so operator skill cannot be separated from the oracle;
and the scaffold itself encodes the core taste, so this design cannot
measure what the oracle would contribute to an unstructured loop.

\section{Discussion}
\label{sec:discussion}

\subsection{What the Three Pillars Buy}

Findings in Section~\ref{sec:findings} share a shape: the loop must
(i)~commit to a falsifiable prediction, (ii)~observe its failure,
and (iii)~choose a non-leaderboard response. The ablation locates
each pillar's contribution to that shape. The card schema and the
content-neutral subagents supply (i) and the floor of (iii) on both
arms: the teacher-bottleneck chain of \S\ref{sec:findings:teacher}
is exactly a non-leaderboard response to two failures, produced with
no oracle in the loop. What kkanbu buys is \emph{which}
non-leaderboard response: the decision to spend an experiment on
test-time adaptation, on a control-framed A/B, or on a categorical
action head tested against a matched control, and the
cross-stream memory that stops the loop from re-buying answers it
already owns (\S\ref{sec:ablation:oracle}).

\subsection{What the Audit Trail Captures, and What It Doesn't}

Every kkanbu consult, brainstormer iteration, card revision, and
promoted tool is persisted on disk by construction; the loop
\emph{needs} these artifacts, they were not retrofitted. Any finding
in Section~\ref{sec:findings} traces to the specific kkanbu turn,
brainstormer iteration, and card that produced it.

The audit trail \emph{does not} capture out-of-band conversations
with collaborators, the interviews that built kkanbu's profile, or
the operator differences between the two arms. A real wet-lab
deployment would need to draw the boundary between the audit-logged
loop and unlogged human context explicitly.

\subsection{Limitations}

Our case study is in simulation, on one domain (quadruped
navigation), with a kkanbu profile seeded by one user; we make no
sim-to-real claim. The ``opinion-only-in-kkanbu'' rule depends on
every other agent respecting its charter; we enforce this through
prompt structure, not provably. The system's output is bounded by
what kkanbu considers good, which sits upstream of any audit trail:
a different researcher's profile would propose different
experiments.

The ablation has two structural limits. First, the scaffold itself
encodes the core of the taste (the mandatory mechanism and
falsification fields, the OOD-weighted composite), so both arms
saturate the taste's presence axes and the design cannot measure
what the oracle would contribute to an unstructured loop; a sharper
ablation needs an arm whose scaffold does not pre-encode the taste.
Second, each arm was run by a different researcher, so arm and
operator are confounded.

Finally, kkanbu audits nothing about its own use. Each call
retrieves fresh from the frozen graph; continuity exists only within
a session, and no mechanism checks a new answer against advice given
in earlier sessions, whether earlier advice was followed, or whether
two streams received contradictory guidance. Its calibration
guardrail is weaker than its framing suggests: of nine auto-flagged
low-confidence answers in the with-kkanbu arm, four were refusals
and five complied anyway with a warning attached, and the run's two
stale-memory slips were caught by the loop's cross-checks, not by
kkanbu. Turning the oracle from an answering service into an
auditor of its own advice is open work.

\section{Conclusion}
\label{sec:conclusion}

We presented an AI Scientist system that separates structural
honesty from research direction. Immutable experiment cards and
content-neutral subagents carry the honesty; kkanbu, a queryable
oracle holding one researcher's taste as a typed knowledge graph,
carries the direction.

We ran the identical loop with and without the oracle across eleven
research streams of a quadruped-navigation benchmark under
obstacle-density shift. Neither arm drifted: both falsified roughly
three quarters of their own hypotheses, obeyed every fair-comparison
rule, and produced mechanism-level findings, including a
representation boundary the two arms exposed jointly and a deployable
inference-time filter that set the round's highest absolute result. The best
trained policy came from the oracle-less arm.

The comparison's lesson is a division of labour. Structure alone
prevents the leaderboard-chasing collapse that motivated this work;
the oracle's contribution is direction: breadth across the
researcher's search axes, authorship of specific winning designs,
and cross-stream memory, at no score premium. The scaffold keeps the
loop honest; kkanbu decides where it looks.

\section*{Impact Statement}

This work proposes structural patterns for AI Scientists that
produce auditable research output. The audit-trail-by-construction
property may help mitigate emerging concerns about attribution and
governance of AI-driven research; the explicit ``opinion-only-in-the-
oracle'' separation makes it easier to ask which decisions came from
the user's stated values and which came from the loop's mechanical
operation. We acknowledge that an AI Scientist of this kind, applied
beyond simulation, would require attention to the boundary between
the logged loop and the unlogged human context that seeds it.

\section*{Acknowledgements}

The computations presented here were conducted in the Resnick High Performance
Computing Center, a facility supported by the Resnick Sustainability Institute
at the California Institute of Technology.


\bibliography{references}
\bibliographystyle{icml2026}

\newpage
\appendix
\onecolumn
\section{The kkanbu Preference Oracle}
\label{app:kkanbu}

This appendix documents the preference oracle used by the AI Scientist
of Section~\ref{sec:system}. The intent is to make precise what kkanbu
\emph{is} (a typed knowledge graph of one user's research taste,
exposed over the Model Context Protocol, MCP), how the profile used in this case study was
\emph{built} (twelve seeded conviction statements via
\texttt{kkanbu\_learn}, then multi-turn \texttt{kkanbu\_curious}
interviews under a privacy-preserving persona protocol), and how it
differs from existing systems for user-specific AI.

\subsection{Philosophy: Research Taste, Not Workflow Preferences}
\label{app:kkanbu_philosophy}

kkanbu is an MCP server~\citep{anthropic2024mcp} that builds and
maintains a persistent typed knowledge graph for a single researcher.
The graph is intended to capture \emph{research taste}: the user's
stance toward what counts as a good question, a sufficient answer, a
real generalisation. It is not for workflow conveniences (preferred
editor, preferred testing framework). In the case-study profile, the
distinction is concrete: of its 54 nodes, 18 are typed \texttt{value},
15 \texttt{pattern}, and 8 \texttt{reasoning}, against a single
\texttt{preference} node. The oracle is asked to speak on the user's
behalf about what counts as good research, not about tool choice.

Three themes recur across the high-confidence value and reasoning
nodes. The first is mechanism over score, held at the profile's
highest confidence: ``\emph{I care about understanding why something
works, not just that it works---a score I can't explain
mechanistically isn't a real result to me}'' ($0.90$). The
mechanism-analyzer's probe turns and card part~6 are answerable to this
stance. The second is invariant competence as the object of study:
``\emph{my research is fundamentally about finding the cheapest
reliable route to inducing the reusable invariant
competence---searching across data, inductive bias, training
objective, and test-time adaptation for the most efficient path}''
($0.92$), refined by a reasoning node holding that the data-vs-method
split ``\emph{isn't a hard line---both data and clever methods can
reach the same invariant competence, so the real question is
efficiency}'' ($0.92$). The OOD-weighted composite of
Section~\ref{sec:problem_setup} is this taste in metric form. The
third is breadth as the denoiser: ``\emph{I deliberately choose a
cheap, sim-only setup with a single clear metric because breadth of
comparable trials is what denoises run-to-run luck and lets an
invariant signal separate from noise}'' ($0.90$). The many-stream
batch design and the three-seed reporting protocol both come from
this node.

These taste statements are what get retrieved when the brainstormer
asks ``which experiment is good?'' The decision rule is not ``which
will score highest'' but ``which one most directly tests whether the
current best policy is \emph{generalising} or \emph{overfitting}.''

Two design choices distinguish kkanbu from a system prompt. First,
\textbf{belief-with-reason}: every node stores a \texttt{content} field
(the stance) and a \texttt{why} field (the reasoning), so retrieval
prompts can reason from stated reasons toward novel questions rather
than pattern-matching surface text. Second, \textbf{first-class
uncertainty}: every node and every answer carries a confidence score,
and low-confidence answers are auto-flagged for review rather than
silently guessed.

\subsection{Profile Schema}
\label{app:kkanbu_schema}

A kkanbu profile is a persistent typed knowledge graph with two
tables. Nodes have a typed payload:
\begin{itemize}
  \item \texttt{node\_type} $\in$ \{\texttt{value}, \texttt{reasoning}, \texttt{pattern}, \texttt{preference}, \texttt{observation}, \texttt{flag}, \texttt{correction}, \texttt{debug\_learning}\}.
  \item \texttt{content} (the stance), \texttt{why} (its reasoning), \texttt{source} (a label recording provenance, including which session type distilled the node), \texttt{confidence} $\in [0, 1]$, and a JSON \texttt{metadata} blob recording further provenance.
\end{itemize}
Edges are typed too: \texttt{relation\_type} $\in$ \{\texttt{supports}, \texttt{derives\_from}, \texttt{refines}, \texttt{generalizes}, \texttt{exemplifies}, \texttt{challenges}, \texttt{contradicts}\}, with a weight and a textual context explaining the relationship. Multiple profiles are isolated per user and selected via \texttt{kkanbu\_of\_who}. The case study uses one profile, the user's research-taste profile.

The case-study profile is a snapshot, frozen before the run, containing 54 nodes and 79 edges. The load-bearing types for the oracle's voice, the value and reasoning nodes, account for $18$ and $8$ entries respectively. Pattern nodes ($n=15$) capture recurring habits of inference; observation nodes ($n=12$) are passive extractions; one preference node records a tool-level preference. Of the 54 nodes, 46 carry confidence $\ge 0.8$. Edge use skews toward argumentative structure: \texttt{supports} (33), \texttt{refines} (12), \texttt{challenges} (11), \texttt{exemplifies} (11), \texttt{derives\_from} (8), \texttt{generalizes} (4).

\subsection{Building the Profile: \texttt{kkanbu\_learn}, \texttt{kkanbu\_curious}, and a Persona Seam}
\label{app:kkanbu_construct}

The case-study profile was built through two intake pathways. (Other
operations exist: \texttt{kkanbu\_ingest} for document extraction,
\texttt{kkanbu\_correct} for human revision, \texttt{kkanbu\_reflect}
for contradiction surfacing. No node in this profile came from
document ingestion; every statement in the graph was either authored
directly by the user or explicitly ratified by them.)

\paragraph{\texttt{kkanbu\_learn}: seeded convictions.} Twelve
de-personalised research convictions were taught explicitly as the
graph's starting position: positions the user could state directly,
such as the single-training-density evaluation stance. Twelve of the
54 nodes carry this provenance.

\paragraph{\texttt{kkanbu\_curious}: Socratic interview at depth.}
The rest of the graph, 42 of 54 nodes ($78\%$, including 18 of the
26 high-confidence value, reasoning, and preference nodes), was
distilled from multi-turn \texttt{kkanbu\_curious} interviews. Given
the current graph, an LLM identifies a knowledge gap and emits a
single question at \emph{value}, \emph{identity}, or \emph{formative}
depth; surface preference questions are rejected by a gatekeeper at
the question-generation step, not merely discouraged in the prompt.
The interview is multi-turn: each answer drills deeper into the same
thread rather than moving on, and once the exchange reaches sufficient
depth the resulting learnings are eagerly distilled into typed nodes
whose \texttt{source} field records the session provenance.

\paragraph{The persona seam: depth without exposure.} The depth
pathway poses a privacy problem. A taste-only profile is too shallow
to extrapolate to unforeseen questions; kkanbu's value is a coherent
\emph{psychology}, not a lookup table. Yet the interviews that
produce this depth push into formative and biographical territory,
and the resulting graph must be shareable: it is a frozen artifact
that both arms of the ablation in
Section~\ref{sec:kkanbu_ablation} consult. The profile therefore
separates three layers. \textbf{Layer 1} (research taste and
judgments) and \textbf{layer 2} (values, what counts as a real
result, what the user hopes the work will become) are the user's
real positions.
\textbf{Layer 3} (formative origin, biography) is a deliberately
fictional but internally coherent persona layer, co-authored during
the interviews and engineered to plausibly generate layers 1--2 while
remaining independent of the user's actual biography. Concretely,
each interview answer was drafted in-persona, then approved or edited
by the user before entering the graph; sessions were audited
afterwards (\texttt{kkanbu\_status}, \texttt{kkanbu\_reflect}) and
any node carrying a real personal fact was scrubbed. All profile
statistics and quotes in this paper draw on layers 1--2.

The reason the loop in Section~\ref{sec:system} consults kkanbu rather than
a system prompt is that the depth pathway has produced statements the
user themselves did not pre-author: they emerged from the interview
and were ratified rather than dictated. A system prompt cannot grow
this way. The snapshot the loop consults was frozen before the streams
launched, and every in-loop call runs with learning disabled, so the
oracle's content is constant across the round; the without-kkanbu
arm's static digest is lifted from the same snapshot.

\subsection{Retrieval and Answering}
\label{app:kkanbu_retrieval}

When an agent calls \texttt{kkanbu\_answer(question, session\_id)},
kkanbu performs the following steps:

\begin{enumerate}
  \item \textbf{Selective retrieval}: keywords are extracted from the question and matched against node content, then the seed set is expanded by breadth-first search (BFS) to depth 2 with score decay, returning a ranked context bundle of nodes and edges.
  \item \textbf{Context assembly}: the bundle is rendered into a structured prompt that separates direct matches from BFS-neighbour context, groups by node type in \emph{identity-first} order (value $\rightarrow$ pattern $\rightarrow$ preference $\rightarrow$ reasoning $\rightarrow$ observation), shows the relationship map between nodes, explicitly tags low-confidence material, and appends a voice directive (``be definitive where confidence is high, hedged where low'').
  \item \textbf{Generation}: an LLM produces a structured response \texttt{\{answer, llm\_confidence, uncertainty\_reasoning\}}.
  \item \textbf{Confidence fusion}: the returned confidence is the lower of graph confidence and LLM self-confidence; below a threshold the answer is automatically annotated with the specific uncertainty reasoning so downstream agents can route the question for human review.
  \item \textbf{Session tracking}: the exchange is appended to the ephemeral session history. On expiry the history is distilled into new graph nodes if it crosses the depth threshold; otherwise it is discarded.
\end{enumerate}

The \texttt{session\_id} parameter is what allows the websearcher and
the brainstormer in Section~\ref{sec:system} to maintain continuity
\emph{within} a search unit or planning group while staying fresh
\emph{across} units: continuity comes from the session, not from the
graph.

\subsection{Relation to Prior Work}
\label{app:kkanbu_related}

kkanbu sits adjacent to several lines of work but does not coincide with any of them.

\textbf{LLM memory frameworks for personalisation.} Mem0~\citep{chhikara2024mem0}, Letta / MemGPT~\citep{packer2023memgpt}, Zep with Graphiti~\citep{rasmussen2025zep}, OpenAI's \emph{Memory} feature, and LangMem store user-related material so that an assistant can recall it. They are typically queried as retrieval-augmented generation (RAG) context inside one assistant's loop, and they extract \emph{facts about the user} (``works at X'') rather than the stances the user holds (values, reasoning patterns). kkanbu's content primitives are typed taste/value/reasoning nodes, and kkanbu is exposed as a separate MCP role that other agents call at decision points, not as memory injected into a single agent's context window.

\textbf{Knowledge-graph-backed agent memory.} GraphRAG / LazyGraphRAG~\citep{edge2024graphrag} and Graphiti model the world's facts (or facts about the user as one entity among many). They have no native distinction between ``fact about the user'' and ``value held by the user.''

\textbf{Preference modelling for LLMs.} RLHF reward models, DPO~\citep{rafailov2023dpo}, and Constitutional AI~\citep{bai2022constitutional} encode preferences, but they are training-time signals baked into model weights, not runtime-queryable artifacts that other agents can address with natural-language questions and inspect as a graph.

\textbf{Cognitive elicitation systems.} GATE~\citep{li2023gate} elicits user preferences via LM-generated questions and produces an unstructured-text profile for one downstream task. Recent work on clarifying-question preference elicitation~\citep{handa2024clarifying} similarly produces single-task latent preference profiles. kkanbu's elicitation (\texttt{kkanbu\_curious}) instead produces a typed graph reused across heterogeneous downstream agents, with depth-gating at the question-generation step and eager distillation of identity-level content into persistent nodes.

\textbf{AI Scientist precedents.} The Sakana AI Scientist~\citep{lu2024aiscientist}, AI-Researcher~\citep{tang2025airesearcher}, Kosmos~\citep{auto2025kosmos}, and similar autonomous research agents include reviewer or critic modules that encode \emph{generic conference taste} or generic novelty/feasibility rubrics. To our knowledge, none expose a per-user research-taste oracle as a distinct queryable agent role.

\textbf{Synthesis.} kkanbu's nearest cousin is Zep/Graphiti, but two deltas are testable: (i) \emph{content}: kkanbu nodes are first-class taste/value/reasoning primitives initialised by Socratic elicitation, not incidentally extracted facts; and (ii) \emph{role}: existing memory systems are RAG context inside one assistant's loop, whereas kkanbu is a separately-addressed oracle other agents call at decision points. We summarise the contribution as: \emph{kkanbu treats the user's research taste as a persistent, queryable graph oracle that other agents in an autonomous-research pipeline must consult before consequential decisions.}

\section{Hyperparameters}
\label{app:hyperparams}

This section provides complete hyperparameter specifications for all methods evaluated in this work. All methods use the Adam optimizer~\citep{kingma2015adam} unless otherwise noted.

\subsection{PIL Control}

Table~\ref{tab:pil_control_params} summarizes the hyperparameters for PIL Control, which directly predicts velocity commands from observations using behavioral cloning. Training uses MSE on velocities:
\begin{equation}
\mathcal{L}_{\mathrm{PIL}}
=
\mathbb{E}_{(o,a^*) \sim \mathcal{D}}
\left[
\|\pi_\theta(o)-a^*\|^2
\right].
\label{eq:pil_loss}
\end{equation}

\begin{table}[h]
\centering
\caption{PIL Control hyperparameters.}
\label{tab:pil_control_params}
\begin{tabular}{ll}
\toprule
\textbf{Parameter} & \textbf{Value} \\
\midrule
Network architecture & MLP [512, 256, 128] \\
Activation & Swish~\citep{ramachandran2017swish} \\
Output activation & Tanh (scaled) \\
Learning rate & $1 \times 10^{-3}$ (cosine) \\
Warmup steps & 3,000 \\
Batch size & 512 \\
Training steps & 600,000 \\
Optimizer & Adam~\citep{kingma2015adam} \\
Early stopping patience & 20 \\
Loss function & MSE \\
\bottomrule
\end{tabular}
\end{table}

\subsection{PIL Trajectory MLP}

Table~\ref{tab:pil_traj_mlp_params} presents hyperparameters for PIL Trajectory MLP, which predicts waypoint sequences processed by a Pure Pursuit controller~\citep{coulter1992purepursuit}. Training uses MSE on the predicted waypoint sequence:
\begin{equation}
\mathcal{L}_{\mathrm{traj}} = \mathbb{E}_{(o,\tau^*)\sim\mathcal{D}}\!\left[\|f_\theta(o)-\tau^*\|^2\right].
\label{eq:traj_loss}
\end{equation}

\begin{table}[h]
\centering
\caption{PIL Trajectory MLP hyperparameters.}
\label{tab:pil_traj_mlp_params}
\begin{tabular}{ll}
\toprule
\textbf{Parameter} & \textbf{Value} \\
\midrule
Network architecture & MLP [512, 256, 128] \\
Activation & Swish~\citep{ramachandran2017swish} \\
Output dimension & 20 (10 waypoints $\times$ 2D) \\
Learning rate & $5 \times 10^{-3}$ (cosine) \\
Warmup steps & 3,000 \\
Min LR ratio & 0.01 \\
Batch size & 1024 \\
Training steps & 600,000 \\
Trajectory scale & 5.0m \\
Lookahead index & 5 \\
\bottomrule
\end{tabular}
\end{table}

\subsection{PIL Trajectory Attention}

Table~\ref{tab:pil_traj_attn_params} details the PIL Trajectory Attention architecture, which incorporates self-attention~\citep{vaswani2017attention} over LiDAR tokens for improved obstacle awareness.

\begin{table}[h]
\centering
\caption{PIL Trajectory Attention hyperparameters.}
\label{tab:pil_traj_attn_params}
\begin{tabular}{ll}
\toprule
\textbf{Parameter} & \textbf{Value} \\
\midrule
Goal branch & 2D $\to$ 64D (2-layer MLP) \\
LiDAR positional encoding & $p_i = i/64$ \\
Token embedding dim & 64 \\
Self-attention heads & 4 \\
Attention output & Sigmoid $\alpha_i \in [0,1]$ \\
Fusion head & MLP [512, 256, 128] over fused 192D \\
Learning rate & $1 \times 10^{-3}$ (cosine) \\
Batch size & 512 \\
Other parameters & As in PIL Traj MLP \\
\bottomrule
\end{tabular}
\end{table}

\subsection{IQL}

Table~\ref{tab:iql_params} presents hyperparameters for Implicit Q-Learning~\citep{kostrikov2022iql}, our offline reinforcement learning baseline. The value function is fit by an asymmetric expectile regression:
\begin{align}
L_V(\psi) &= \mathbb{E}_\mathcal{D}\!\left[L_\tau^2(Q_\theta(s,a)-V_\psi(s))\right], \label{eq:iql_v}\\
L_\tau^2(u) &=|\tau-\mathbf{1}(u<0)|\,u^2, \label{eq:iql_tau}
\end{align}
and the policy is extracted by AWR
\begin{equation}
\mathcal{L}_\pi(\phi) = -\mathbb{E}_\mathcal{D}\!\left[e^{\beta\,\mathrm{clip}(A(s,a),0,c)}\log\pi_\phi(a\mid s)\right].
\label{eq:iql_awr}
\end{equation}

\begin{table}[h]
\centering
\caption{IQL hyperparameters.}
\label{tab:iql_params}
\begin{tabular}{ll}
\toprule
\textbf{Parameter} & \textbf{Value} \\
\midrule
Actor network & MLP [256, 256] \\
Critic network & MLP [256, 256, 256] (double Q) \\
Value network & MLP [256, 256, 256] \\
Activation & ReLU \\
Learning rate & $3 \times 10^{-4}$ (cosine) \\
Warmup steps & 1,000 \\
Batch size & 256 \\
Training steps & 600,000 \\
Expectile $\tau$ & 0.7 \\
Temperature $\beta$ & 3.0 \\
Discount $\gamma$ & 0.99 \\
Target update rate & 0.005 \\
AWR clip range & $[0, 5.0]$ \\
AWR max weight & 20.0 \\
Dataset size & ${\sim}8.65$M transitions \\
\bottomrule
\end{tabular}
\end{table}

\paragraph{Reward function.}
\label{app:reward}
The reward used during offline-RL training  combines a sparse goal bonus, a dense distance-reduction term, and a path-following term against the same A*~\citep{hart1968astar} expert that produced the demonstrations:
\begin{equation}
R(s, a, s') = R_{\text{sparse}} + R_{\text{dense}} + R_{\text{path}},
\end{equation}
\begin{align}
R_{\text{sparse}} &= \begin{cases} 10.0 & \text{if } d(s') \leq 0.51\,\text{m} \\ 0 & \text{otherwise} \end{cases} \\
R_{\text{dense}} &= 0.5 \cdot (d(s) - d(s')) \\
R_{\text{path}} &= 5.0 \cdot \exp(-d_{\text{waypoint}})
\end{align}
where $d(s)$ is the distance to goal and $d_{\text{waypoint}}$ is the distance to the nearest A* waypoint.

\subsection{PPO}

Table~\ref{tab:ppo_params} summarizes hyperparameters for Proximal Policy Optimization~\citep{schulman2017ppo}, our online reinforcement learning baseline. Value function estimation uses Generalized Advantage Estimation~\citep{schulman2016gae}.

\begin{table}[h]
\centering
\caption{PPO hyperparameters and initial reward terms.}
\label{tab:ppo_params}
\begin{tabular}{ll}
\toprule
\textbf{Parameter} & \textbf{Value} \\
\midrule
Policy network & MLP [512, 256, 128] \\
Value network & MLP [512, 256, 128] \\
Learning rate & $3 \times 10^{-4}$ \\
Clipping $\epsilon$ & 0.15 \\
Entropy coefficient & 0.003 \\
Discount $\gamma$ & 0.997 \\
GAE $\lambda$ & 0.97 \\
Parallel environments & 512 \\
Unroll length & 25 \\
Updates per batch & 8 \\
Total timesteps & $5\times 10^8$ \\
\midrule
\multicolumn{2}{c}{\textbf{Initial Reward Terms}} \\
\midrule
Goal completion reward & $+250.0$ \\
Goal progress reward & $+50.0 \cdot \Delta d$ \\
Collision penalty & $-10.0$ \\
Fall penalty & $-50.0$ \\
Velocity smoothness penalty & $-1.0 \|c_t-c_{t-1}\|^2$ \\
Backward velocity penalty & varies by experiment \\
Time-to-collision penalty & varies by experiment \\
Stuck-detection penalty & varies by experiment \\
Waypoint progress reward (privileged PPO only) & varies by experiment \\
\bottomrule
\end{tabular}
\end{table}

\subsection{Pure Pursuit Controller}

Table~\ref{tab:pure_pursuit_params} documents the Pure Pursuit controller~\citep{coulter1992purepursuit} parameters used for trajectory-based methods.

\begin{table}[h]
\centering
\caption{Pure Pursuit controller parameters.}
\label{tab:pure_pursuit_params}
\begin{tabular}{ll}
\toprule
\textbf{Parameter} & \textbf{Value} \\
\midrule
Lookahead index & 5 \\
Trajectory horizon & 10 waypoints \\
Trajectory scale & 5.0m \\
Max forward velocity & 0.75 m/s \\
Max angular velocity & 0.75 rad/s \\
Proportional gain $k_p$ & 1.0 \\
Slowdown distance & $0.5$\,m (= goal threshold) \\
Alignment minimum (MAF) & $0.0$ (deployment); $0.3$ for the always-forward expert variant \\
\bottomrule
\end{tabular}
\end{table}

\input{figures/appendix_experiments.tex}

\section{Implementation Details}
\label{app:implementation}

\subsection{Environment Specification}

Table~\ref{tab:env_params} summarizes the simulation environment parameters. All experiments use the environment implemented in MuJoCo~\citep{todorov2012mujoco} with GPU acceleration via the MJX framework~\citep{freeman2021brax}. We use LiDAR rather than RGB because MJX supports efficient ray-based sensing in JAX while parallel RGB rendering is unsupported. Episodes terminate on goal reach (within $0.5\,\mathrm{m}$), collision, fall, or timeout.

\begin{table}[h]
\centering
\caption{Environment parameters.}
\label{tab:env_params}
\begin{tabular}{ll}
\toprule
\textbf{Parameter} & \textbf{Value} \\
\midrule
Environment & \texttt{Go1NavigationForestEnvV5} \\
Training room & 30m $\times$ 30m \\
Control frequency & 50 Hz \\
Episode length & 3,000 steps (60s) \\
Goal threshold & 0.5m \\
Robot radius & 0.3m \\
Obstacle density & 0.04 trees/m$^2$ \\
Obstacle radius & 0.3m \\
LiDAR beams & 64 \\
LiDAR range & 10m \\
$v_x$ range & $[0,\,1.5]\,\mathrm{m/s}$ \\
$v_y$ range & $[-0.6,\,0.6]\,\mathrm{m/s}$ \\
$v_{\mathrm{yaw}}$ range & $[-1.2,\,1.2]\,\mathrm{rad/s}$ \\
\bottomrule
\end{tabular}
\end{table}

\subsection{Observation Space}

Table~\ref{tab:obs_space} details the observation space components (75D total). The encoding deliberately omits absolute position so the policy is scale-invariant to room size.

\begin{table}[h]
\centering
\caption{Observation space components (75D total).}
\label{tab:obs_space}
\begin{tabular}{lcp{6cm}}
\toprule
\textbf{Component} & \textbf{Dims} & \textbf{Description} \\
\midrule
Normalized goal distance & 1 & $\mathrm{clip}(d_{\mathrm{goal}}/D_0,\,0,\,1)$, $D_0 = 30\sqrt{2}$\,m \\
Normalized heading-to-goal & 1 & Heading-to-goal difference, mapped to $[-1, 1]$ \\
LiDAR & 64 & Normalized 360$^\circ$ ring, $[0,1]$ \\
Action history & 9 & Flattened previous-3 velocity commands $(v_x, v_y, v_{\mathrm{yaw}})$ \\
\bottomrule
\end{tabular}
\end{table}

\subsection{Data Collection}

Expert demonstrations are collected using A*~\citep{hart1968astar} planning on an inflated occupancy grid (0.2m resolution) with Pure Pursuit~\citep{coulter1992purepursuit} tracking (3.0m lookahead). The expert has access to the complete obstacle map: privileged information unavailable to the student during deployment. We collect ${\sim}8.65$M expert transitions, shared by the PIL and offline-RL paradigms.

\subsection{Computational Resources}

Experiments ran on a shared cluster, predominantly on NVIDIA H200 and L40S GPUs, using the JAX/Flax framework with MJX simulation. PIL methods train in under 2 hours, IQL requires approximately 4 hours, and PPO requires approximately 6 hours with 512 parallel environments.


\section{The Program Description}
\label{app:program}

Both arms read the same program description at every step: it is the
one document every subagent shares, and in the without-kkanbu arm it
is the only source of direction. We reproduce it below in full, as
the agents read it. Local context is masked as
\texttt{<...>}: cluster-specific filesystem paths, the frozen
locomotion checkpoint identifier, and the oracle profile's install
path. Internal codenames are kept so the document stays authentic;
the glossary: \emph{Round~5} is the run this paper reports;
\emph{R1--R4} are earlier development runs whose numerical findings
the document excludes by rule; \emph{round5b} is the without-kkanbu
arm; \emph{d06--d11} are the six discovery streams (Hybrid~I/II,
New-paradigm~I/II, Tuning~I/II of Appendix~\ref{app:experiments});
\emph{ADR~$n$} cites an internal architecture-decision record;
\emph{V5} is the fifth revision of the observation space (the
document's \S13.4 gives the lineage); SLURM is the cluster's job
scheduler.

Two passages are superseded by how the ablation actually ran. The
document's \S1 and \S7 lock a \emph{planned} without-oracle arm in
which a static digest of the taste profile is pasted into the
subagent prompts. That digest was never built: the arm as run
consulted no taste artifact at all
(Section~\ref{sec:kkanbu_ablation}), so the measured contrast is
live oracle versus no oracle, not live retrieval versus a static
digest of the same taste.

\lstset{basicstyle=\ttfamily\scriptsize, breaklines=true,
  columns=fullflexible, keepspaces=true, upquote=true,
  breakindent=0pt}
\begin{multicols}{2}
\lstinputlisting{figures/program_round5_masked.txt}
\end{multicols}

\end{document}

%% file: figures/top3_cross_round.tex

\begin{table*}[t]
\centering
\small
\setlength{\tabcolsep}{4pt}
\caption{Leading verified results per ablation arm. All entries are
3-seed (mean $\pm$ stratified percentile bootstrap 95\% half-width;
anti-conservative at $n{=}3$: under a Student-$t$ interval some of
these beats become inconclusive). Each $\Delta$ is against that row's
own reference (the stream's anchor for paradigm rows; the designated
anchor or champion reference for tuning and hybrid rows), so rows are
not mutually comparable. The collision-cone entry is an
\emph{inference-time} result on frozen weights and is not comparable
to trained-policy rows; its honest figure is the within-model delta.
$^{\dagger}$best-of-$N$ selection on the scoring densities (operating
point chosen post hoc; the interval does not cover selection). DAgger
denotes imitation learning with iterative on-policy expert
corrections~\citep{ross2011dagger}.}
\label{tab:top3_cross_round}
\begin{tabular}{@{}lp{6.0cm}llr@{}}
\toprule
Arm & Entry & Class & Composite & $\Delta$ vs.\ reference \\
\midrule
\textbf{Without kkanbu}
 & imitation student on regenerated evasive-expert demos (hybrid stream, batch 3) & trained & $90.96 \pm 1.52$ & $+12.24$ \\
 & privileged PPO, goal-progress reward rescale (batch 2) & trained & $88.80 \pm 1.12$ & $+2.97$ \\
 & privileged PPO, potential-based shaping (batch 1) & trained & $86.56 \pm 2.42$ & $+0.73$ \\
\midrule
\textbf{With kkanbu}
 & DAgger correction-depth sweep, 16 rounds (tuning stream, batch 4)$^{\dagger}$ & trained & $88.48 \pm 2.44$ & $+1.20$ (tie) \\
 & DAgger rollout-mixture sweep, $\beta{=}0.5$, on the champion recipe (tuning stream, batch 2) & trained & $87.96 \pm 0.83$ & $+0.68$ (tie) \\
 & DAgger distillation student, control arm (hybrid stream, batch 1) & trained & $87.28 \pm 0.56$ & $+1.45$ \\
\cmidrule(l){2-5}
 & collision-cone filter on the frozen DAgger student$^{\dagger}$ & inference-time & $92.43 \pm 0.69$ & $+4.89$ within-model \\
\bottomrule
\end{tabular}
\end{table*}

%% file: figures/appendix_experiments.tex
\section{Per-Experiment Results}
\label{app:experiments}

This appendix lists every experiment the paper reports, one row per
experiment, ordered by stream and batch so each stream's progression
reads top to bottom. The inclusion rule is the frozen list of
Section~\ref{sec:ablation}: an experiment appears only if its
evaluation is complete at three seeds; still-running work is excluded
(the jobs continue for future rounds). Composites are the winning arm
of each experiment as mean $\pm$ population standard deviation over
the three seed composites, recomputed from the evaluation records on
disk; sweeps report the winning arm, with three-seed sibling arms
noted in the description. The two hybrid, two new-paradigm, and two
tuning streams are numbered I and II. Both new-paradigm streams of
the with-kkanbu arm are absent: every result they completed is a
single-seed falsified terminal under the round's falsified-discovery
exception, so none passes the three-seed rule (the process statistics
of Section~\ref{sec:ablation:honesty} still count them).

\subsection*{With kkanbu}

{\small
\begin{longtable}{@{}p{2.3cm}lp{5.2cm}llp{2.6cm}@{}}
\caption{With-kkanbu arm: all three-seed-complete experiments.}
\label{tab:appendix_5a}\\
\toprule
Stream & Batch & Experiment (single mutated axis) & Composite & $\Delta$ vs.\ ref. & Verdict \\
\midrule
\endfirsthead
\toprule
Stream & Batch & Experiment (single mutated axis) & Composite & $\Delta$ vs.\ ref. & Verdict \\
\midrule
\endhead
\bottomrule
\endlastfoot
PPO (anchor 83.0) & 1 & raise the PPO entropy cost, single knob & $83.44 \pm 1.22$ & $+0.44$ & falsified (null) \\
 & 2 & cone-gated inverse-TTC braking reward, coefficient sweep & $85.33 \pm 0.87$ & $+2.33$ & falsified (per-density legs) \\
\midrule
Privileged PPO (anchor 85.83) & 1 & critic-only nearest-obstacle clearance descriptor & $86.65 \pm 1.84$ & $+0.82$ & inconclusive \\
 & 2 & one cone-gated braking reward term, coefficient sweep & $87.09 \pm 1.59$ & $+1.26$ & inconclusive (tie) \\
\midrule
IQL (anchor 62.98) & 1 & actor ported onto the standard core, attention demoted to auxiliary & $61.46 \pm 3.52$ & $-1.52$ & falsified (null) \\
 & 2 & reward-scale sweep with a scale-invariant checkpoint filter & $65.78 \pm 2.66$ & $+2.80^{c}$ & inconclusive \\
 & 3 & advantage-weighting temperature sweep at matched training steps & $63.33 \pm 3.82$ & $+0.35$ & falsified (flat) \\
\midrule
PIL Control (anchor 78.72) & 1 & Gaussian input-noise augmentation of the demonstrations & $79.96 \pm 2.19$ & $+1.24$ & inconclusive \\
 & 2 & corrective-recovery label synthesis, dataset-only change & $79.13 \pm 1.09$ & $+0.41$ & falsified (null) \\
 & 3 & categorical action head, paired $+4.54$ over a freshly trained matched MSE control ($81.33 \pm 0.63$) & $85.87 \pm 0.14$ & $+7.15$ & not falsified \\
\midrule
PIL Trajectory (anchor 63.39) & 1 & add one step of action history & $63.54 \pm 1.24$ & $+0.15$ & falsified (flat) \\
 & 3 & trained lateral nudge head after the tracker; $+4.09$ of the gain is attributable against a matched fixed-lookahead control & $78.48 \pm 0.85$ & $+15.09$ & not falsified \\
 & 4 & focal reweighting of the nudge loss, exponent sweep (zero recovers batch 3) & $78.11 \pm 0.93$ & $-0.37$ vs.\ batch 3 & falsified (monotone harm) \\
\midrule
Hybrid I (ref.\ 83.0) & 1 & set-invariant vs.\ fixed-vector LiDAR encoder A/B & $79.20 \pm 1.22$ & $-3.80$ & falsified (both arms below) \\
\midrule
Hybrid II (champion ref.$^{d}$) & 1 & privileged-teacher DAgger distillation, A/B of a density-invariance auxiliary head; the auxiliary arm falsified ($75.74 \pm 1.71$) & $87.28 \pm 0.51$ & (is the ref.) & control arm = co-leader \\
 & 2 & scene-family holdout generalisation diagnostic (in-family $86.80 \pm 0.66$) & $87.33 \pm 2.17$ & tied & diagnostic \\
 & 3 & collision-cone velocity filter on the frozen batch-1 student, gate sweep (filter off: $87.54 \pm 0.76$) & $92.43 \pm 0.61$ & $+4.89$ w.m.$^{a}$ & inference class \\
\midrule
Tuning I (ref.\ 83.0, then champion$^{d}$) & 1 & raise the PPO entropy cost & $84.80 \pm 1.06$ & $+1.80$ & falsified (clause) \\
 & 2 & closing-speed proximity reward, coefficient sweep & $84.30 \pm 0.35$ & $+1.30$ & falsified \\
 & 3 & smooth inverse-TTC reward form, coefficient sweep & $83.26 \pm 0.85$ & $+0.26$ & falsified \\
 & 4 & correction-depth sweep of the champion distillation recipe & $88.48 \pm 2.14$ & $+1.20$ (tie) & falsified (tail bar) \\
\midrule
Tuning II (ref.\ 85.83, then champion$^{d}$) & 1 & shorten the discount horizon, single knob & $65.48 \pm 12.91$ & $-20.35$ & falsified (unstable) \\
 & 2 & rollout-mixture sweep of the champion distillation recipe & $87.96 \pm 0.69$ & $+0.68$ (tie) & falsified (tail) \\
 & 3 & clearance-gated relabelling of teacher commands, cap sweep & $87.09 \pm 0.43$ & $-0.19$ (tie) & falsified \\
 & 4 & clearance-gated speed cap on the frozen champion, gate sweep (cap off: $87.20 \pm 0.39$) & $87.70 \pm 0.09$ & $+0.50$ w.m.$^{a}$ & inference class, null \\
\end{longtable}
}

\subsection*{Without kkanbu}

{\small
\begin{longtable}{@{}p{2.3cm}lp{5.2cm}llp{2.6cm}@{}}
\caption{Without-kkanbu arm: all three-seed-complete experiments.}
\label{tab:appendix_5b}\\
\toprule
Stream & Batch & Experiment (single mutated axis) & Composite & $\Delta$ vs.\ ref. & Verdict \\
\midrule
\endfirsthead
\toprule
Stream & Batch & Experiment (single mutated axis) & Composite & $\Delta$ vs.\ ref. & Verdict \\
\midrule
\endhead
\bottomrule
\endlastfoot
PPO (anchor 83.0) & 1 & training-only LiDAR observation noise & $81.91 \pm 0.82$ & $-1.09$ & falsified \\
 & 2 & distance-gated braking reward & $83.26 \pm 0.85$ & $+0.26$ & falsified (null) \\
 & 3 & TTC-gated braking reward & $85.63 \pm 0.19$ & $+2.63$ & falsified (missed the pre-registered bar) \\
\midrule
Privileged PPO (anchor 85.83) & 1 & critic-only nearest-obstacle clearance descriptor & $86.56 \pm 1.99$ & $+0.73$ & inconclusive \\
 & 2 & double the goal-progress shaping coefficient & $88.80 \pm 1.12$ & $+2.97$ & not falsified (all seeds clear) \\
\midrule
IQL (anchor 62.98) & 1 & actor onto the standard core; raise the expectile & $63.39 \pm 3.38$ & $+0.41$ & falsified (null) \\
\midrule
PIL Control (anchor 78.72) & 1 & beam-dropout and range-jitter augmentation & $81.11 \pm 2.26$ & $+2.39$ & inconclusive \\
 & 2 & same recipe, longer training, checkpoint chosen by closed-loop composite & $83.67 \pm 0.72$ & $+4.95^{b}$ & not falsified \\
\midrule
PIL Trajectory (anchor 63.39) & 1 & multimodal conditional-VAE waypoint head & $53.54 \pm 0.47$ & $-9.85$ & falsified \\
\midrule
Hybrid I (ref.\ 85.83) & 1 & permutation-invariant set-encoder auxiliary on the privileged recipe & $85.33 \pm 2.27$ & $-0.50$ & falsified (bar) \\
 & 2 & add a braking reward to the batch-1 recipe & $83.00 \pm 3.79$ & $-2.83$ & falsified (all legs) \\
\midrule
Hybrid II (ref.\ 78.72) & 1 & density-perturbation augmentation with a consistency loss & $73.96 \pm 0.48$ & $-4.76$ & falsified (desensitisation) \\
 & 3 & evasive strafe-and-brake layer on the expert; demonstrations regenerated, same student retrained & $90.96 \pm 1.52$ & $+12.24$ & not falsified (champion) \\
\midrule
New-paradigm I (anchorless, retention clause$^{e}$) & 2 & density-equivariant attention-over-beams encoder & $72.33 \pm 9.10$ & $R = 0.821$ & not falsified \\
 & 4 & global dilated circular-CNN encoder, globality-vs-attention cut & $71.15 \pm 9.04$ & $R = 0.759$ & falsified (marginal) \\
\midrule
New-paradigm II (anchorless, paired A/B) & 1 & density-shift view augmentation with an output-consistency loss, paired against an unaugmented control ($78.81 \pm 1.70$) & $80.80 \pm 0.41$ & $+1.98$ paired & inconclusive \\
 & 2 & open-space proximity gate on the consistency term, paired & $80.72 \pm 3.19$ & $+1.91$ paired & falsified (collision leg) \\
\midrule
Tuning I (ref.\ 83.0) & 1 & raise the PPO entropy cost & $84.96 \pm 2.29$ & $+1.96$ & falsified (clause) \\
\midrule
Tuning II (ref.\ 78.72) & 1 & LiDAR observation-noise knob on the control-paradigm recipe & $79.11 \pm 0.88$ & $+0.39$ & falsified (null) \\
\end{longtable}
}

\footnotesize
$^{a}$~Inference-time results on frozen weights are a separate class
and are not rank-comparable with trained rows; the honest figure is
the within-model (w.m.) delta against the filter-off control.
$^{b}$~Checkpoint selected by composite on the scoring densities; the
audit quantifies the selection optimism at about $1.9$ points.
$^{c}$~The winning reward-scale arm trained for more steps than its
siblings; the gain is step-confounded and overlaps the anchor CI.
$^{d}$~Hybrid and tuning batches reference the cross-stream champion
(the batch-1 DAgger control arm, $87.28$) once it existed; earlier
batches reference their paradigm anchor. ``Tie'' marks CI-overlapping
results.
$^{e}$~This stream pre-registered an OOD-retention ratio $R$ (success
on the out-of-distribution densities relative to in-distribution
success) as its metric; its falsification bar is on $R$, not the
composite.
\normalsize